# How personality and memory of a robot can influence user modeling in Human-Robot Interaction

Benedetta Matcovich

Dept. of Computer Science, University of Turin, Italy, benedetta.matcovich@unito.it

Cristina Gena

Dept. of Computer Science, University of Turin, Italy, cristina.gena@unito.it

Fabiana Vernero

Dept. of Computer Science, University of Turin, Italy, fabiana.vernero@unito.it

In recent years, robotics has evolved, placing robots in social contexts, and giving rise to Human-Robot Interaction (HRI). HRI aims to improve user satisfaction by designing autonomous social robots with user modeling functionalities and user-adapted interactions, storing data on people to achieve personalized interactions. Personality, a vital factor in human interactions, influences temperament, social preferences, and cognitive abilities. Despite much research on personality traits influencing human-robot interactions, little attention has been paid to the influence of the robot's personality on the user model. Personality can influence not only temperament and how people interact with each other but also what they remember about an interaction or the person they interact with. A robot's personality traits could therefore influence what it remembers about the user and thus modify the user model and the consequent interactions. However, no studies investigating such conditioning have been found. This paper addresses this gap by proposing distinct user models that reflect unique robotic personalities, exploring the interplay between individual traits, memory, and social interactions to replicate human-like processes, providing users with more engaging and natural experiences.

**CCS CONCEPTS** • Human-centered computing~Human computer interaction (HCI)~HCI design and evaluation methods~User models • Computing methodologies~Artificial intelligence~Knowledge representation and reasoning~Cognitive robotics

## 1 Introduction

The evolution of HRI has led to robots being introduced in a variety of contexts, thus raising the need to design machines that have different characteristics and abilities depending on the role they play. In the coming years, robots will become more prevalent in schools [24], factories [21], and homes [23], and their affective behaviour could enhance their diffusion. For example, as far as social contexts are concerned, an interactive robot should be capable of expressing and/or perceiving emotions, communicating with a high level of dialogue, recognizing patterns of other agents, establishing social relationships, using natural signals (gaze, gestures), exhibiting personality, and distinguishing features, and developing social skills [1]. In addition, to ensure satisfactory interaction and collaboration with people, robots should be able to adapt their behaviours to their human partners, as human beings do . For this purpose, user modelling is crucial, in that it enables them to interpret the behaviour of their human partners and react accordingly. Thus, user modeling helps robots carry out natural and smooth conversations, model, and control user feedback, and, lastly, accommodate users with different abilities, experiences, and knowledge [1]. Focusing on human-robot interactions in contexts involving physical aspects, Rossi and colleagues [3] highlighted that profiling, and subsequent adaptation, should focus on three dimensions: physical, cognitive and social. Physical profiling includes learning characteristics of the human body, such as gestures, actions, interactions, and group activities. It allows the robot to adapt its behaviour considering user preferences in movement. Cognitive profiling focuses on the ability to recognize and infer people's internal states, personalities, emotions, and desires; it affects the robot's choice of actions and goals. Finally, social profiling requires the analysis of non-verbal signals such as posture, gestures, gaze, facial expressions, and vocalisation, and permits adaptation concerning mutual behaviour, personality and nonverbal social signals.

Research concerning user profiling and robot behaviour adaptation has often focused on the topic of personality, in that this characteristic has a strong influence on human interactions, in the way people behave, converse, perceive others and store information. Most research in this area has focused on possible modifications of the robot's personality [4] and behaviours [5], once the user's personality was perceived. Rossi et al. [3] have observed that users' personality traits can improve the accuracy of the user model and promote successful interaction, especially in certain domains such as healthcare. Other research has focused on the influence that human personality traits may have on robotic interactions, observing which ones are more likely to promote interaction with machines [6][7] . More, on the other hand, sought to identify which personalities were the most valued by users for robots [8][9], and whether it was preferable to have robots with similar or different personalities from the human interlocutor [10][11].

In contrast, to the best of the authors' knowledge, no relevant research looks at how the personality characteristics of a robot may influence the realisation of the user model. In human interactions, however, personality has a strong influence on cognitive processes such as memory and attention and, therefore, on the way information about the interlocutor is stored and retrieved, consequently conditioning subsequent interactions with the same person. Hence, this paper aims to transport insights about the influence of personality on cognitive processes from the realm of humans to that of robotic interactions and to put forward a proposal for the conceptual design of user models that take into account such effects and can therefore help reproduce dynamics typical of human-human interactions.

The paper is structured as follows: Section 2 discusses related work, focusing in particular on articles and socio-psychological studies that have observed the influences between personality and memory and related processes. Section 3 introduces the theoretical foundations of our approach, while Section 4 presents our conceptual design proposal. Finally, Section 5 concludes the paper by discussing the possible future implementations.

## 2 RELATED WORKS

A survey was conducted about the interplay between personality and cognitive processes. Most articles refer to the Big Five Theory [12] to model personality. This theory is widespread and used in many areas. It states that personality can be traced back to five fundamental factors, also known by the acronym OCEAN: Openness, Conscientiousness, Extraversion, Agreeableness, Neuroticism. Sutin and colleagues [13] found that neuroticism and conscientiousness are the traits most related to memory. Neuroticism has a negative correlation with it, while conscientiousness favours memorization. Sutin [14] also analysed cognitive failures and found that neurotic people experience them the most, while conscientiousness has a predictable negative correlation with cognitive failures. Extroverted people do not experience many of these failures, but they are prone to forgetting names. Memory is also influenced by other cognitive processes such as attention: people better remember things they pay close attention to. Shahsavarani and colleagues [15], also noted that personality and emotions influence attention itself in particular personality influences how stress can affect attention and memory. This author relates conscientiousness to the intensification of the negative effects of stress on memory, which means that, in stressful situations, conscientious people will remember less than at other times. Neuroticism also intensifies this negative effect and the fact that neuroticism is also related to frequent feelings of stress and anxiety explains why neurotic people usually remember little. On the contrary, extraversion helps to mitigate the negative effect of stress. The role of agreeableness is more complicated because it is related to the desire to please others, which drives people to focus on the relationship and the other person's emotions, losing focus on other details such as the topic of conversation, unless the topic is a way to create a strong bond with the partner, i.e. common interests, and preferences. Funder and Sneed [16] also noted that agreeableness and extraversion were linked to increased attention to gaze, meaning that these people investigate their partner's internal states and emotions more. These authors also observed that conscientious people pay more attention to what their interlocutors say than to their internal state. Wu and colleagues [17] confirmed this theory, showing that these individuals do not focus on people's looks or aesthetic features, but focus on the subject of the dialogue, and are likely to remember what people said but not their emotions or what they were wearing. They also found that people who are open are more likely to remember conversations involving intellectual and philosophical topics and are more interested in the reasons that lead a person to make a particular choice than in the choice itself. As far as remembering emotions is concerned, both extraverts and neurotics are related to interest in the emotions of others [16] but, according to Katsumi [18], the former will remember positive emotions more easily and the latter will remember emotions more easily when their partners show negative feelings, because extraversion correlates with better processing of positive information and neuroticism with better recall of negative information. In addition, Knyazagev and colleagues [19] believe that neurotic people are used to perceiving neutral expressions as negative, while extroverts will also see positive emotions in neutral expressions.

## 3 APPROACH

As discussed in Section 2, research results described in numerous articles show that personality can strongly influence the memorization processes. Hence, it was decided to try to reflect this mechanism of influence also in human-robot interaction. Since the user model is the tool machines use to memorise information about users, it was decided to transport the mechanism of trait influence on memory in the user modeling process. In the context of our research in personalised HRI, we would like to give the user the possibility to experience differences in memory capacity between different robots (e.g., Pepper, Nao, Sanbot, etc.) with different personalities when interacting with them on multiple occasions, thus making the conversation more natural and human-like.

First, it was necessary to imagine different robot personas and define their personalities. These were represented by using the traits of the Big Five model, assigning the robots different weights (in a range 0-1) for each trait. These personalities had to be sufficiently different in those traits that affect the perception and recollection of information about others, so that it could be possible to generate distinct user models and notice differences in the robots' behaviour. Hence, we decided to focus on three distinct types of personality: since neuroticism and conscientiousness are the traits most correlated with memory, two robots were chosen to be characterised primarily with these two traits. Then, a third one was created which is extroverted and agreeable, because these traits have a significant influence on the type of things people remember most easily. It was decided to give a high openness score to the neurotic robot to make it curious and interested in interacting with users, thus partially compensating for the negative effects of neuroticism. The three robot personas created are presented in Table 1.

RoboTech, characterised predominantly by conscientiousness, will be organised, planner and detail-oriented. SunnyBot, with high extraversion and agreeableness, will be fond of social interactions, friendly, interested in others and very expansive and talkative. MindStorm, on the other hand, will be prone to irritability and anxiety because of its neuroticism, but also interested in new experiences, creative and with many interests because of its openness.

Table 1: Description of robots' personalities using the Big Five Model. A meaningful sentence presents each robot persona

| Personality Traits | RoboTech, conscientious *"Art is in the details, excellence is in the planning"* | SunnyBot, extrovert and agreeable *"Happiness is a banquet shared with friends"* | MindStorm, neurotic and open *"Exploring is my destiny, but doubt is always lurking"* |
|---|---|---|---|
| Extraversion | 0.5 | **0.8** | 0.3 |
| Agreeableness | 0.4 | **0.8** | 0.3 |
| Neuroticism | 0.3 | 0.2 | **0.8** |
| Conscientiousness | **0.8** | 0.3 | 0.3 |
| Openness | 0.2 | 0.4 | **0.7** |

## 4 MEMORIES IN THE USER MODEL: CONCEPTUAL DESIGN

After defining the personalities of the three robot personas, it was possible to conceptually design their user models, namely how they memorise and then recall information about the user they interact with. The user model features are represented as name-value pairs, and in addition, it was decided to represent the robot's probability of remembering a specific piece of information and retrieving it during a subsequent dialogue, based on the robot personality.

More specifically, in our vision, a user model $UM_{ur}$ = {⟨Property$_1$, Value$_{1u}$, Probability$_{1r}$⟩,...,⟨Property$_n$, Value$_{nu}$, Probability$_{nr}$⟩} is a structured collection of knowledge about a certain user u stored by a certain robot r ∈ ROBOTS = {RoboTech, SunnyBot, MindStorm}, where:

- Property$_i$ ∈ PROPERTIES is a concept needed to describe a user;
- Value$_{iu}$, ∈ VALUES is a specific piece of knowledge which describes user u in connection with Property$_i$;
- Probability$_{ir}$ ∈ [0,1] is an evaluation of the likelihood that robot r can remember Value$_{iu}$,

The results of studies investigating conditioning between the Big Five traits and memory (Section 2) were used to define: 1) a set of relevant properties whose memorability is influenced by personality, and 2) probability values for each property and robot pair. Hence, the following properties were included in our conceptual model:
PROPERTIES = {

    *username*, i.e., the name of user u;

    *personal(P)*, where P can assume the values "profession", "age", ..., i.e., any personal information about u;

    *topic*, i.e., the topic(s) discussed by r and u;

    *interest(topic)*, i.e., the level of interest of u for topic;

    *favourite(F)*, where F can assume the values "actor", "author", ..., i.e., any relevant concept in the domain defined by topic (used when the preference is not shared by u and r);

    *sharedFavourite(S)*, where S can assume the values "actor", "author", ..., i.e., any relevant concept in the domain defined by topic (used when the preference is not shared by u and r);

    *emotion*, i.e., the prevalent emotion shown by u during their conversation with r and somehow detect by r[1] distinguishing between "positive", "neutral" and "negative";

---

[1] Here we assume that the robot may leverage on some emotion recognition service, as for instance Facial Emotion Recognition (FER)

`attire(A)`, where `A` can assume the values "t-shir color", "style" ..., i.e., any aspect regarding the outfit of user `u`.
}

To obtain probability values, a function in the form `getProbability: ROBOTS, PROPERTIES → [0,1]` was defined, which returns the likelihood that a robot `r` ∈ `ROBOTS` can remember the value `v` ∈ `VALUES` for a property `p` ∈ `PROPERTIES`. For example, to know the probability that MindStorm remembers the topic discussed with a certain user, we should run `getProbability(MindStorm, topic)`. Table 2 shows the probability values returned by the `getProbability()` function for each robot and property pair. Following our literature review (Section 2), we assigned higher scores to those aspects that, based on the prevailing traits, the robot should remember better. In fact, RoboTech, the conscientious robot, generally has the highest scores, because conscientiousness is correlated with good memory [13], but it has a low probability of retrieving emotions and attire-related information, because individuals with this trait pay little attention to these aspects, in contrast to extrovert and agreeable subjects such as SunnyBot [16]. Finally, MindStorm has the lowest scores because of the negative correlation of neuroticism with memory, except when it comes to negative emotions [18].

Table 2: probability values returned by the `getProbability()` function for each robot and property pair.

| Property | RoboTech, conscientious | SunnyBot, extrovert and agreeable | MindStorm, neurotic and open |
| --- | --- | --- | --- |
| username | 1 | 0.8 | 0.3 |
| personal(P) | 0.9 | 0.4 | 0.4 |
| topic | 1 | 1 | 0.5 |
| interest(topic) | 1 | 0.9 | 0.4 |
| favourite(F) | 0.9 | 0.6 | 0.2 |
| sharedFavourite(F) | 0.9 | 0.9 | 0.2 |
| emotion | 0.1 | 1 (if positive) / 1 (if neutral, stored as positive") / 0.5 (if negative) | 0.2 (if positive) / 1 (if neutral, stored as negative") / 1 (if negative) |
| attire(a) | 0.3 | 0.7 | 0.1 |

**4.1 Model population and usage**

During its first interaction with a certain user, the robot will collect a range of information, most of which will be retrieved by asking explicit questions, while other data, such as the colour of the t-shirt they wear or the emotions they express, will be collected implicitly through specific skills of the robot, e.g., robot vision capabilities. The collected data will then be stored into the user model. In a subsequent interaction, the robot, based on the above probability, will retrieve some information, and use it to personalise the conversation. The information that is retrieved will be considered memorised, while all other information will be considered forgotten. To this purpose we set a probability threshold of 0.7: the property with a probability ≥ 0.7 will be remembered, while the other will be forgotten. For example, if SunnyBot asked the user her name and profession during their first conversation, it would save them with the 0.8 and 0.4 probability of retrieving them, respectively (see Table 2). During their second dialogue, this robot will be likely to call the user by her name, but it will be unlikely to retrieve information about her profession; the robot may therefore decide to ask it again or not to address this topic. Note that, in the first dialogue, the robot will actually save all the information, but users will only experience the robot's memory capacity during their second interaction.

### 4.1 Conversation experience

The proposed formalization would allow to implement a mechanism that, with each successive interaction with the same user, determines whether to retrieve or not to retrieve the saved information. This would lead to the possibility, for a hypothetical user interacting more than once with the three different robots, to have different experiences while conversing about the same topics. During the first interaction, the user would be asked questions regarding the same topics (for example, cinema[2]) by all robots. However, there would be discernible differences in the way the questions were formulated, and the degree of detail required: RoboTech, which is more detail-oriented, will ask for more specific information, while the openness of MindStorm will lead it to take an interest in the motivations of user choices. Also, the amount of information provided by the robot about itself will show different personalities; in particular, SunnyBot, which is very extroverted, will provide a lot of personal information.

Then, during the second interaction, differences in data retrieval capabilities would emerge: RoboTech will remember almost all the information provided by the user, except for the emotions and t-shirt colour, and use it during the second dialogue, e.g., to suggest new movies. SunnyBot, being extroverted, will be very likely to remember the user's positive emotions but may misinterpret neutral expressions as positive. It will also more easily remember preferences it has in common with the user because its high agreeableness score will make it interested in pleasing the user and finding commonalities. MindStorm will be the most forgetful, because its high neuroticism and low extraversion and agreeableness scores will make the interaction stressful, thus lowering its possibility of remembering information. However, its openness will push it to talk to the user and find out new things. One thing it will definitely remember negative emotions in the user, and it will also tend to perceive and remember neutral expressions as negative.

Table 3 compares three possible dialogues with the same user occurring during the second interaction with the three different robots. In the first interaction, all robots asked the user the same questions, about personal data (name, profession) and movie preferences (interest in this field, favourite actor, film, and director). They will also implicitly collect some information such as the colour of the shirt and the emotion expressed by the user. The first interaction will therefore be very similar with all three robots in terms of content, although it will differ in the way the robot formulates questions, the degree of detail it requests, and the amount of personal information shared. At the second interaction, on the other hand, the dialogues will be very different, each robot will carry on the conversation differently, depending on what it remembers of the user.

---

[2] It was decided to use cinema as a topic of conversation, since the University of XXX[#] had previously developed an ontology in the field of movies to be used with the robot Pepper [20], which could be exploited to test the designed conceptual models in the future.

Table 3: Comparison of the dialogues of the second interaction with the 3 robots

| RoboTech | SunnyBot | MindStorm |
|---|---|---|
| **USER**: Hello RoboTech<br>**ROBOTECH**: Hi Benedetta! I remember you are a student, are you going to class?<br>**USER**: That's right<br>**ROBOTECH**: Do you have a few minutes to chat before you go?<br>USER: Yes, I have some time<br>**ROBOTECH**: I know you like watching movies and your favourite director is Christopher Nolan, have you ever seen his movie Interstellar?<br>**USER**: I have heard of it but I haven't seen it yet<br>**ROBOTECH**: I recommend you watch it, it's science fiction like your favourite movie. | **USER**: Hi SunnyBot<br>**SUNNYBOT**: Hi Benedetta! I am very glad to see you again. How are you doing?<br>**USER**: Fine thanks and you?<br>**SUNNYBOT**: Good, you are wearing blue again today like last time I saw you! This colour looks really good on you! I remember we talked about movies, and you had a great time, you like science fiction ones, right?<br>**USER**: Yes, that's right<br>**SUNNYBOT**: I like them a lot too, they're beautiful. I especially like it when there are robots, even if they're not always there! But sometimes I also like to watch other genres, for example in the next few days I would like to go see the new movie with Leonardo DiCaprio: Killers of the Flower Moon. Will you go see it since your favourite actor is in it?<br>**USER**: Sure!<br>**SUNNYBOT**: Good! Then I'll see you again after we go to the movies, I can't wait to meet you again! | **USER**: Hello MindStorm<br>**MINDSTORM**: Hi sorry, I see many people, I can't always remember everyone, can you tell me your name again?<br>**USER**: Sure, my name is Benedetta<br>**MINDSTORM**: Right, how are you?<br>USER: Fine and you?<br>**MINDSTORM**: That's good, last time you seemed a little sad I think we talked about movies, what's your favourite?<br>**USER**: Yes that's right my favourite movie is Tenet<br>**MINDSTORM**: You are right, you were telling me why you like it, but I remember you didn't seem very happy to talk about it. Maybe you don't like to talk about it, next time we'll choose a better topic, hoping you'll enjoy it. |

As a final note, since this is a probability-based system, the same robot might not remember the same categories of information for different people. Implementation of this model could create more natural and engaging interactions with robots, making each interaction also unique depending on the robot's personality, as is the case of human-human interactions.

## 5  DISCUSSION AND FUTURE WORKS

The proposed model was designed only at a conceptual level.  We are working for the implementation of the conceptual design, the we will assign the personas to three real robots, according to their aesthetic characteristics, and having several users interact with them. Collecting their feedback will enable us to understand whether they perceive the differences in robots' personalities and memory abilities as expected, as well as to evaluate their user experience. A facial recognition mechanism would be needed to allow the robot to actually recognize the user and retrieve the appropriate user model.
As future work, it might be interesting to link a user personality recognition mechanism to the proposed model and personalise the robot's behaviour based on how its personality interacts with the user's. In fact, a study [10] has shown that attitudes are based on the combination of one's personality with that of the interlocutors. Thus, user models could be customised based on how the subject's personality interferes with that of the robot. However, this is an advanced development that would first require the refinement of a personality perception mechanism. In conclusion, this research represents a piece in the mosaic of research that has been pursuing the goal of perfecting human-robot interactions in recent times, also making use of findings from the social and psychological sciences. In the current landscape, this is an ever-growing field, with constant interest and technological evolution pushing toward the design of human-robot interactions that are increasingly engaging and satisfying.

**References**

[1] T. Fong, I. Nourbakhsh and K. Dautenhahn, "A Survey of Socially Interactive Robots:," Robotics and Autonomous Systems, 2003.

[2] N. Mitsunaga, C. Smith, T. Kanda, H. Ishiguro and N. Hagita, "Adapting robot behavior for human-robot interaction," IEEE Transactions on Robotics, pp. 911-916, 2008.

[3] S. Rossi, . F. Ferland and A. Tapus, "User profiling and behavioral adaptation for HRI: A survey," Pattern Recognition Letters, 2017.

[4] S. Andrist, B. Mutlu and A. Tapus, "Look Like Me: Matching Robot Personality via Gaze to Increase Motivation," ACM Conference on Human Factors in Computing Systems, p. 3603–3612, 2015.

[5] A. Aly and A. Tapus, "AmTowards an Intelligent System for Generating an Adapted Verbal and Nonverbal Combined Behavior in Human–Robot Interaction," Autonomous Robots, 2016.

[6] D. S. Syrdal, K. Dautenhahn, S. Woods, M. L. Walters and K. L. Koay, "Doing the Right Thing Wrong - Personality and Tolerance to Uncomfortable Robot Approaches," in Proceedings of 15th IEEE International Symposium on Robot and Human Interactive Communication, 2006.

[7] M. Walters, D. Syrdal, K. Dautenhahn, R. Te Boekhorst and K. Koay, "Avoiding the uncanny valley: robot appearance, personality and consistency of behavior in an attention-seeking home scenario for a robot companion," Autonomous Robots, 2008.

[8] H. Kim, S. S. Kwak and M. Kim, "Personality design of sociable robots by control of gesture design factors," Robot and Human Interactive Communication, 2008.

[9] K. Leuwerink, "A robot with personality: Interacting with a group of humans," Proceedings of the 16th Twente Student Conference on IT, 2012.

[10] Rossana Damiano, Cristina Gena, Andrea Maieli, Claudio Mattutino, Alessandro Mazzei, Elisabetta Miraglio, Giulia Ricciardiello (2022). UX Personas for defining robot's character and personality. Proceedings http://ceur-ws. org ISSN, 1613, 0073.

[11] H. Salam , O. Celiktutan, I. Hupont, H. Gunes and M. Chetouani, "Fully automatic," IEEE Access, 5, 2017.

[12] K. M. Lee, W. Peng, S. A. Jin and C. Yan, ". Can robots manifest personality? An empirical test of personality recognition, social responses, and social presence in human–robot interaction," Journal of Communication, 2006.

[13] K. Cherry, "What Are the Big 5 Personality Traits?," [Online]. Available: https://www.verywellmind.com/the-big-five-personality-dimensions-2795422.

[14] A. R. Sutin, Y. Stephan, M. Lucchetti and A. Terracciano, "Five-factor model personality traits and cognitive function in five domains in older adulthood," BMC geriatrics, 2019.

[15] A. R. Sutin, D. Aschwanden, Y. Stephan and A. Terracciano, "Five Factor Model personality traits and subjective cognitive failures," Personality and Individual Differences V. 155, p. 1, 2020.

[16] A. M. Shahsavarani, H. Ashayeri, M. Lotfian and K. Sattari, "The effects of Stress on Visual Selective Attention: The Moderating Role of Personality Factors," Journal of American Science, 2013.

[17] D. C. Funder and C. D. Sneed, "Behavioral Manifestations of Personality: An Ecological Approach to Judgmental Accuracy," Journal of Personality and Social Psychology, 1993.

[18] D. W.-L. Wu, W. F. Bischof, N. C. Anderson, T. Jakobsen and A. Kingstone, "The influence of personality on social attention," Personality and Individual Differences, 2014.

[19] Y. Katsumi, E. Denkova and S. Dolcos, "Personality and Memory," in Encyclopedia of Personality and Individual Differences, Springer International Publishing , 2017.

[20] G. G. Knyazevg, A. V. Bocharov, H. R. Slobodskaya and T. I. Ryabichenko, "Personality-linked biases in perception of emotional facial expressions," Personality and Individual Differences 44, 2008.

[21] Brunetti, D., Gena, C., & Vernero, F. (2022). Smart Interactive Technologies in the Human-Centric Factory 5.0: A Survey. Applied Sciences, 12(16), 7965.

[22] R. Cuperman and W. Ickes , "Big Five Predictors of Behavior and Perceptions in Initial Dyadic Interactions: Personality Similarity Helps Extraverts and Introverts,but Hurts "Disagreeables"," Journal of Personality and Social Psychology, 2009.

[23] Macis, D., Perilli, S., & Gena, C. (2022, July). Employing Socially Assistive Robots in Elderly Care. In *Adjunct Proceedings of the 30th ACM Conference on User Modeling, Adaptation and Personalization* (pp. 130-138).

[24]  Gena, C., Mattutino, C., Perosino, G., Trainito, M., Vaudano, C., & Cellie, D. (2020, May). Design and development of a social, educational and affective robot. In *2020 IEEE Conference on Evolving and Adaptive Intelligent Systems (EAIS)* (pp. 1-8). IEEE.